\documentclass[letterpaper, 10pt, journal, twoside]{IEEEtran}

\usepackage{caption}
\usepackage{subcaption}
\usepackage{graphics}
\usepackage{amsmath}
\usepackage{amssymb} 
\usepackage[linesnumbered,ruled,vlined]{algorithm2e}
\usepackage[dvipsnames*,svgnames,x11names]{xcolor}
\usepackage{setspace}
\usepackage{makecell}
\usepackage{graphicx}
\usepackage[utf8]{inputenc}
\DeclareUnicodeCharacter{2217}{\textasteriskcentered}
\usepackage{booktabs}
\usepackage{hyperref}
\hypersetup{hidelinks}
\urldef{\codeurl}\url{https://brsrikrishna.github.io/BBoE_RAL/}
\usepackage[capitalize]{cleveref}
\usepackage{booktabs,makecell,tabularx}
\renewcommand{\arraystretch}{1.0}

\usepackage{graphicx,adjustbox,booktabs,makecell}
\usepackage[final]{changes}
\definechangesauthor[name={Spot Red}, color=red]{spot}
\SetAlFnt{\scriptsize}
\newcommand{\ava}[1]{{\color{Black}ava:#1} }
\newcommand{\ar}[1]{{\color{ForestGreen}\textbf{AR:#1}} }
\newcommand{\hg}[1]{\textcolor{orange}{[HG: #1]}}

\newcommand{\revdel}[1]{{\color{red}#1}}

\title{
BBoE: Leveraging Bundle of Edges for Kinodynamic Bidirectional Motion Planning 
}
\usepackage{booktabs,makecell}

\author{Srikrishna Bangalore Raghu and Alessandro Roncone%
\thanks{Manuscript received: 2 July 2026, Revised: 19 September 2026, Accepted: 14 October 2026.}%
\thanks{This paper was recommended for publication by Editor Lucia Pallottino upon evaluation of the Associate Editor and Reviewers’ comments.}%
\thanks{The authors are with the Department of Computer Science, University of Colorado Boulder, Boulder, CO, USA (Email: {\tt\footnotesize srba2850@colorado.edu}). Code and supplementary material are available at \codeurl.}%
\thanks{Digital Object Identifier (DOI): see top of this page.}}%

\renewcommand{\revdel}[1]{}
\renewcommand{\ar}[1]{}
\renewcommand{\ava}[1]{}

\renewcommand{\hg}[1]{}

\begin{document}

\maketitle

\begin{abstract}
In this work, we introduce \textsl{BBoE}, a bidirectional, kinodynamic, sampling-based motion planner that consistently and quickly finds solutions in environments with varying obstacle clutter. The algorithm combines exploration and exploitation while relying on precomputed robot motions, resulting in efficient convergence towards the goal. Our key contributions include: i) a strategy to quickly find solutions in obstacle-rich spaces by leveraging a bundle of precomputed robot motions (i.e., edges); and ii) BBoE, a robust bidirectional kinodynamic planner that utilizes this strategy to produce fast and feasible solutions. 
Across four systems spanning 3D to 6D state spaces, BBoE is the only planner to achieve $\geq96\%$ success on every system while attaining the lowest planning time on three of four.
\end{abstract}

\section{INTRODUCTION}
\IEEEPARstart{A}{s} robots are increasingly being deployed around us, they must operate in obstacle-rich scenarios (i.e., drones navigating disaster sites, mobile robots threading through a crowd of humans, and vehicles maneuvering in dense traffic), where they need to quickly find feasible and collision-free paths. This falls under the problem of motion planning: identifying a series of valid configurations that a robot can assume to transition from an initial state to a desired goal state. Among the popular techniques, sampling-based motion planning is a graph-based approach used to generate robot motions by sampling discrete states and establishing connections between them via edges \cite{Review}.  
Traditionally, these techniques employ a unidirectional tree that grows from the start state and expands towards the goal region 
\cite{RRT, thomason2024motions}. 
A faster alternative uses a bidirectional approach, where two trees rooted at the start and goal converge towards each other \cite{RRTCONNECT,GBRRT}. In either case, these planners need to account for the constraints on the robots' kinematics (position, velocity) and dynamics (acceleration, forces, torques), also known as \textsl{kinodynamic constraints} 
\cite{Kinodynamic_Kuffner}.
\begin{figure}[t]
\centering
\includegraphics[width=\linewidth]{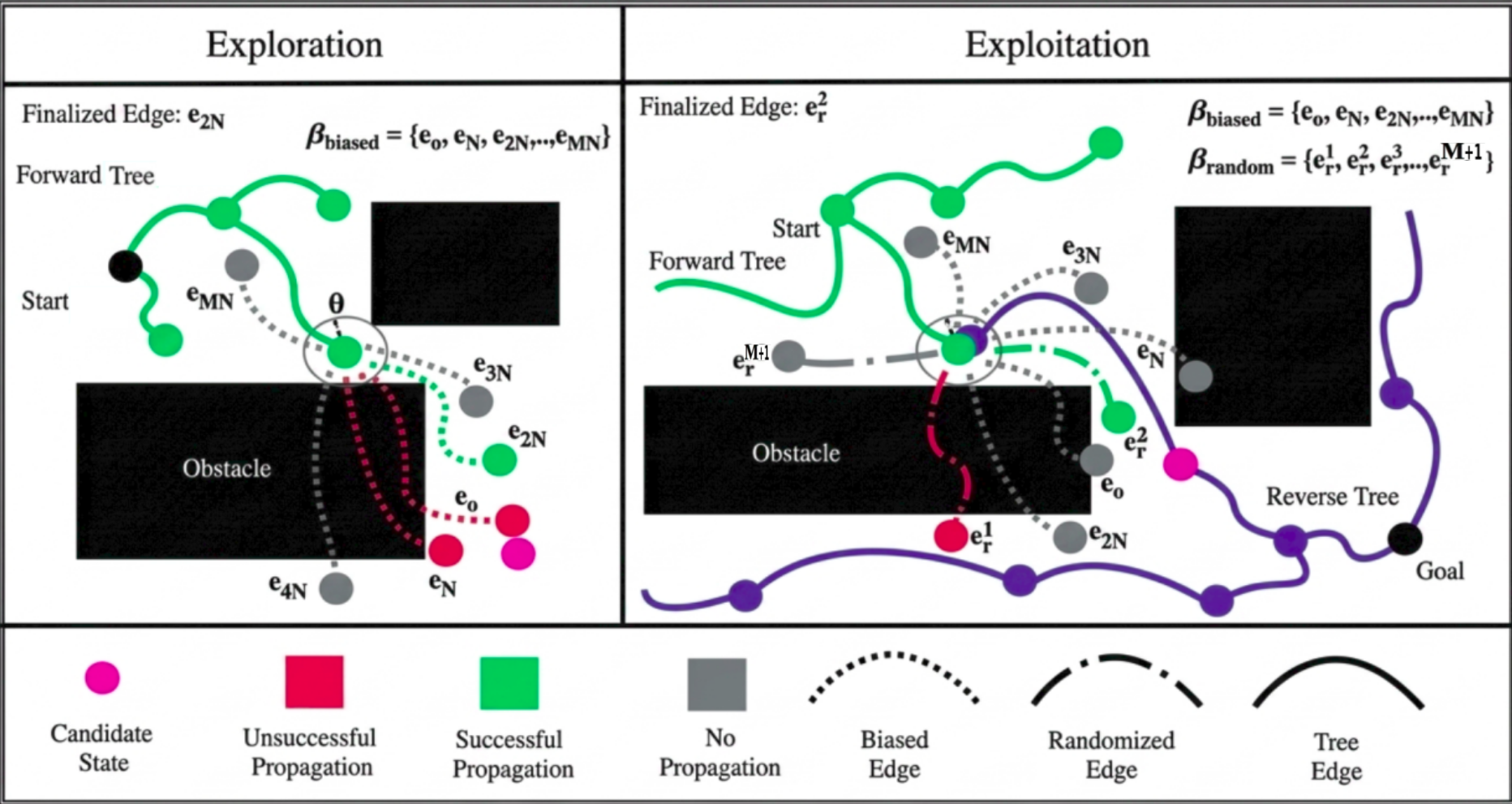}
\centering
\caption{The functionality of BBoE spans two phases, exploration and exploitation. During exploration, the forward tree primarily uses the biased edges ($\boldsymbol{\beta}_{\text{\upshape biased}}$) to expand towards the candidate state, while the reverse tree relies on propagating random edges backward in time. During exploitation, either the biased edges or the randomized edges ($\boldsymbol{\beta}_{\text{\upshape random}}$) are used while allowing the forward tree to accurately trace the reverse tree towards the goal. 
}
\label{fig:architecture}
\vspace{-5pt}
\end{figure}

Generating motions that respect these constraints is inherently difficult: unlike geometric planning, where any two nearby states can be connected by a straight line, kinodynamic systems can only move in ways their dynamics permit. This becomes especially challenging as the number of obstacles in the environment increases, resulting in very few possible solutions and the consequently low probability of finding them.
A common approach for kinodynamic motion generation is forward propagation, which executes random control inputs for random durations in a Monte-Carlo attempt to reach a desired state \cite{GBRRT, monte}. In cluttered spaces, it results in high planning times when accurate connections are preferred, high-cost paths when the connections are less accurate, and low success rates within a stipulated time limit. The Bundle of Edges (BoE) algorithm \cite{BoE} removes this online cost by precomputing a large set of feasible motions (i.e., edges) offline and reusing them as a roadmap. However, BoE expands its tree by propagating every edge near the current state, resulting in repeated obstacle collisions and wasted propagation effort.

Thus, whether motions are generated online or drawn from a precomputed bundle, current kinodynamic methods cannot reliably find solutions in highly cluttered and obstacle-dense environments without sacrificing speed. Overcoming this barrier requires a planner that reuses precomputed motions to avoid online propagation cost while handling them intelligently so that scarce solutions are found quickly even as clutter increases. Motivated by this, we present the Bidirectional Bundle of Edges (BBoE) algorithm which makes precomputed-motion planning viable in heavy clutter by reliably finding dynamically feasible solutions within practical time budgets across systems ranging from ground vehicles to aerial robots. This entails two contributions:
\begin{itemize}
  \item A strategy to quickly find solutions in obstacle-rich spaces by sorting and sequencing a bundle of precomputed robot motions (i.e., edges).
  \item A bidirectional sampling-based motion planner that leverages this strategy to provide feasible solutions with reduced runtime and a high success rate amidst varying obstacle clutter, including higher-dimensional systems where existing precomputed-motion planners fail.
\end{itemize}

\section{BACKGROUND AND RELATED WORK}
\subsection{Kinodynamic Motion Planning}
To generate robot motions that follow kinematic and dynamic constraints, kinodynamic planners use two methods: Steering Functions and Forward Propagation. Steering functions use robot dynamics to analytically or numerically connect pairwise states, and have been used by mainstream planners like RRT \cite{Kinodynamic_Kuffner} and RRT* \cite{steering_needed_frazzoli}. However, they are computationally expensive and challenging to define for complex systems \cite{steering_rrt*_berg, needs_steering}. To address the complexities of steering functions, planners such as SST \cite{sst} and DIRT \cite{dirt} use forward propagation, a simpler and cheaper alternative. These planners sample several control inputs per expansion and keep the propagation ending closest to the desired state, thus treating the dynamics as a black box. This black-box treatment also extends forward propagation to contact-rich tasks, e.g., non-prehensile poking \cite{pasricha2022pokerrt}. An alternative is to precompute robot motions offline and reuse them online. Motion primitive (MP) methods build a library of short, dynamically feasible trajectories that define an implicit graph over reachable states. These are usually constructed lazily during an A*-like search with edges ranked by a heuristic \cite{mp2}. 
A related line of work, iDb-RRT \cite{idb_rrt} relaxes exact connections by allowing a bounded discontinuity, repaired afterward by trajectory optimization. BoE~\cite{BoE}, LazyBoE~\cite{LazyBoE}, KiTE-Extend~\cite{gupta2026efficientmultirobotmotionplanning}, and NPM-LMJ \cite{briscoe2024} apply the precomputed-motion idea in the sampling-based setting without discontinuities, using a bundle of edges as a probabilistic roadmap \cite{PRM}.
\\However, these methods do not focus on quickly finding solutions amidst a clutter of obstacles. Planners that have tried to tackle this contain drawbacks. RRT*-Smart \cite{clutter_rrt*} uses RRT* to generate an initial plan while relying on steering functions, while OracleNet \cite{clutter_learning} requires training for every new environment to find narrow pathways between states. 
Spill-free liquid transport in clutter \cite{abderezaei2024clutter} plans a geometric path with RRT* and enforces dynamics only afterwards through time parameterization, while CAT-RRT \cite{nechyporenko2023catrrt} traverses dense clutter by admitting contact, which the robot and its surroundings must tolerate.
Furthermore, these algorithms only comprise unidirectional trees, which are slower at finding solutions compared to bidirectional alternatives.
\subsection{Bidirectional Motion Planning}
Bidirectionality in sampling-based motion planning was introduced by RRT-Connect \cite{RRTCONNECT}, where two trees grow towards each other until they're connected, resulting in lower planning times. This benefit is seen in subsequent bidirectional planners, such as AIT* \cite{ait} and RRT*-Connect \cite{Review}. Bidirectional searches have previously attempted to quickly find solutions in cluttered spaces, but they contain drawbacks. B-Spline RRT \cite{bidirectional_clutter_car} is tailor-made specifically for cars, while  B$\text{\upshape I}^\text{\upshape 2}$RRT* \cite{bidirectional_clutter_propagation} banks on basic online propagation for traversal.
 In the kinodynamic domain, these algorithms traditionally depend on steering functions to connect the trees. Conversely, the Generalized Bidirectional Rapidly Exploring Random Tree (GBRRT) \cite{GBRRT} avoids the complex computation of connecting the trees by enabling the reverse tree to guide the forward tree towards the goal. This spans over two phases. During the exploration phase, both trees explore the state space until they're close enough. The exploitation phase allows the forward tree to use the reverse tree as a heuristic and advance towards the goal. However, both phases rely on online Monte-Carlo propagations while targeting desired states, resulting in significant computation time and low success rates in heavily cluttered environments. Next, we present how BBoE combines precomputed edge handling with bidirectional search to enable fast navigation through obstacle clutter.
\section{METHODS}
\subsection{Problem Formulation}
\label{sec:problem}
In the kinodynamic domain, the state space comprises the robot's configuration (e.g., positions and angles) alongside its higher-order derivatives (e.g., velocities and accelerations). We define the robot state space $Q \subset \mathbb{R}^m$ containing $m$-dimensional states $x\in Q$ alongside the control space $U \subset \mathbb{R}^n$ consisting of $n$-dimensional controls $u\in U$. Given time $t$, the forward dynamics model of the robot is defined by:\vspace{5 pt}
\centerline{$\dot{x} = f(x(t),u(t))$~~~~~  $x\in Q$, $u\in U$}
\\
\\
Given a start state $x_0$ and a goal region $Q_{goal} \subseteq Q$, BBoE solves the \textsl{feasibility} problem: find a collision-free, dynamically feasible trajectory $\pi$,\\

\centerline{$\pi(t) : [0,1] \rightarrow Q_{free} \subseteq Q$,~~  $\pi(0)= x_0$,~$\pi(1) \in Q_{goal}$,} where every segment of $\pi$ results from integrating $f$ under an admissible control. We measure a planner by its time to the first solution and by its success rate within a fixed budget. BBoE claims no optimality with respect to path cost. Like GBRRT \cite{GBRRT}, BBoE grows a forward tree from $x_0$ and a reverse tree from $Q_{goal}$ in two phases. During \textsl{exploration}, both trees expand toward randomly sampled states until they arrive within a threshold distance of each other. During \textsl{exploitation}, the forward tree expands toward reverse-tree nodes, which act as a heuristic leading it into the goal region.

\subsection{Bundle of Edges} 
To enable accurate propagation from one state to the other, we leverage a bundle of precomputed robot motions (i.e., a bundle of edges) similar to the BoE \cite{BoE} planner. An edge bundle consists of an expansive number of forward propagations randomly generated in the robot's state space. The bundle is computed offline and stored for future use. Once a motion planning query is initiated, the bundle is retrieved, and the edges that initiate in the vicinity of the robot's current state are used as a reference/map to guide the robot to the desired state. The computation performed offline reduces the burden on the planner during the online (i.e., planning) phase.\\
The edge bundle $\varepsilon$ is formed when a set of random start states $x_0 \in Q$ undergoes forward propagation by applying random control inputs $u \in U$ for random time durations $\Delta t$, resulting in the robot reaching final states $x_f\in Q$:\\
\\
\centerline{$\varepsilon$ = $\{(x_0,u,\Delta t, x_f)|x_0,x_f\in Q, u\in U \}$}\\
\\where $x_f = x(\Delta t)$ is the solution of $\dot{x}=f(x,u)$ with $x(0)=x_0$ under the edge's control $u$ over $[0,\Delta t]$. We write a generic edge as $\text{e} = (x_0, u, \Delta t, x_f) \in \varepsilon$. Subscripts (e.g., $\text{e}_i$) are used only after the edges are sorted in \cref{sec:strategy}, where $i$ denotes the position in the sorted list.
\begin{figure}[t]
\centering
\subfloat[Randomized Edges\label{fig:strategy-a}]{
  \includegraphics[width= \linewidth]{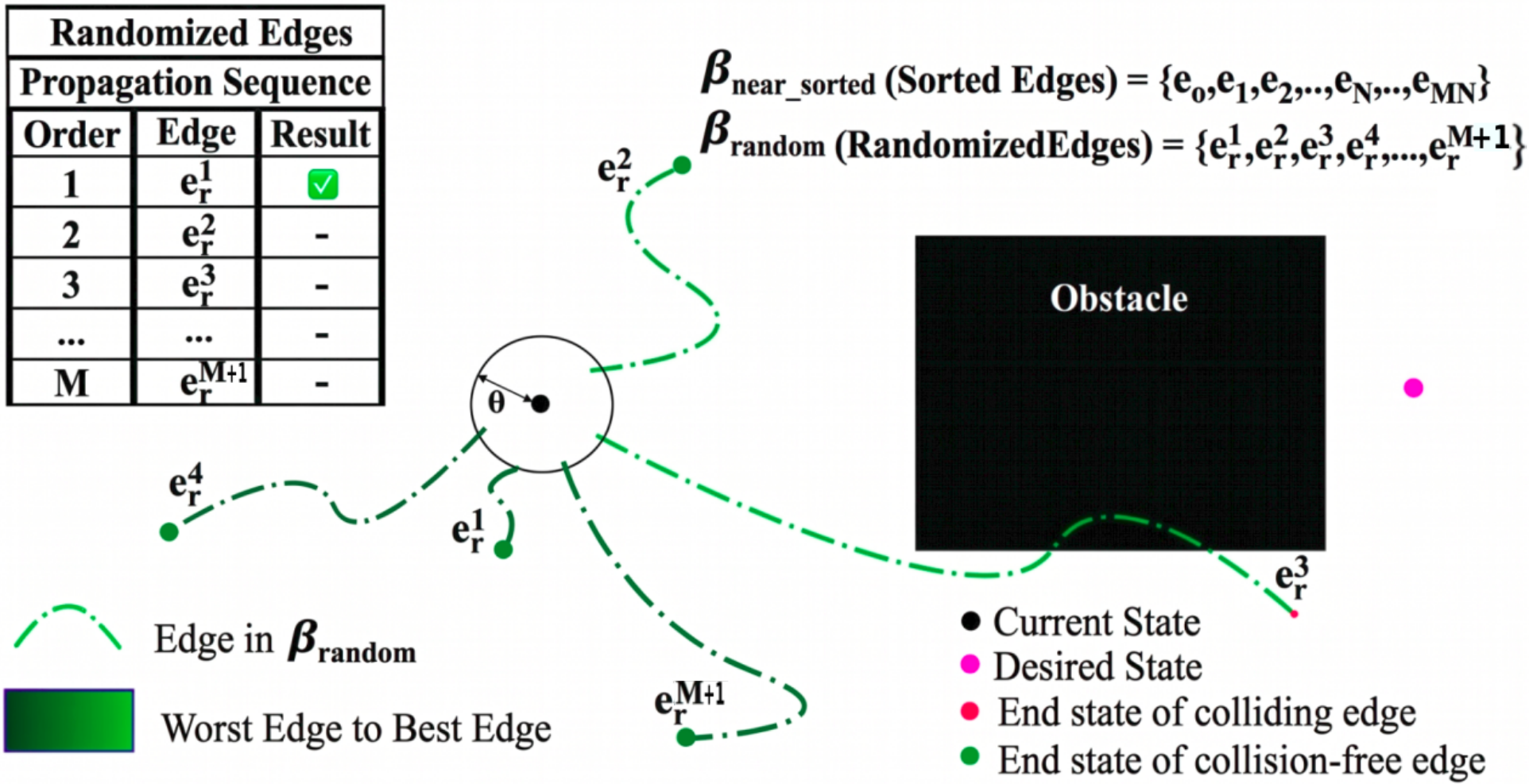}
}

\subfloat[Biased Edges\label{fig:strategy-b}]{
  \includegraphics[width=\linewidth]{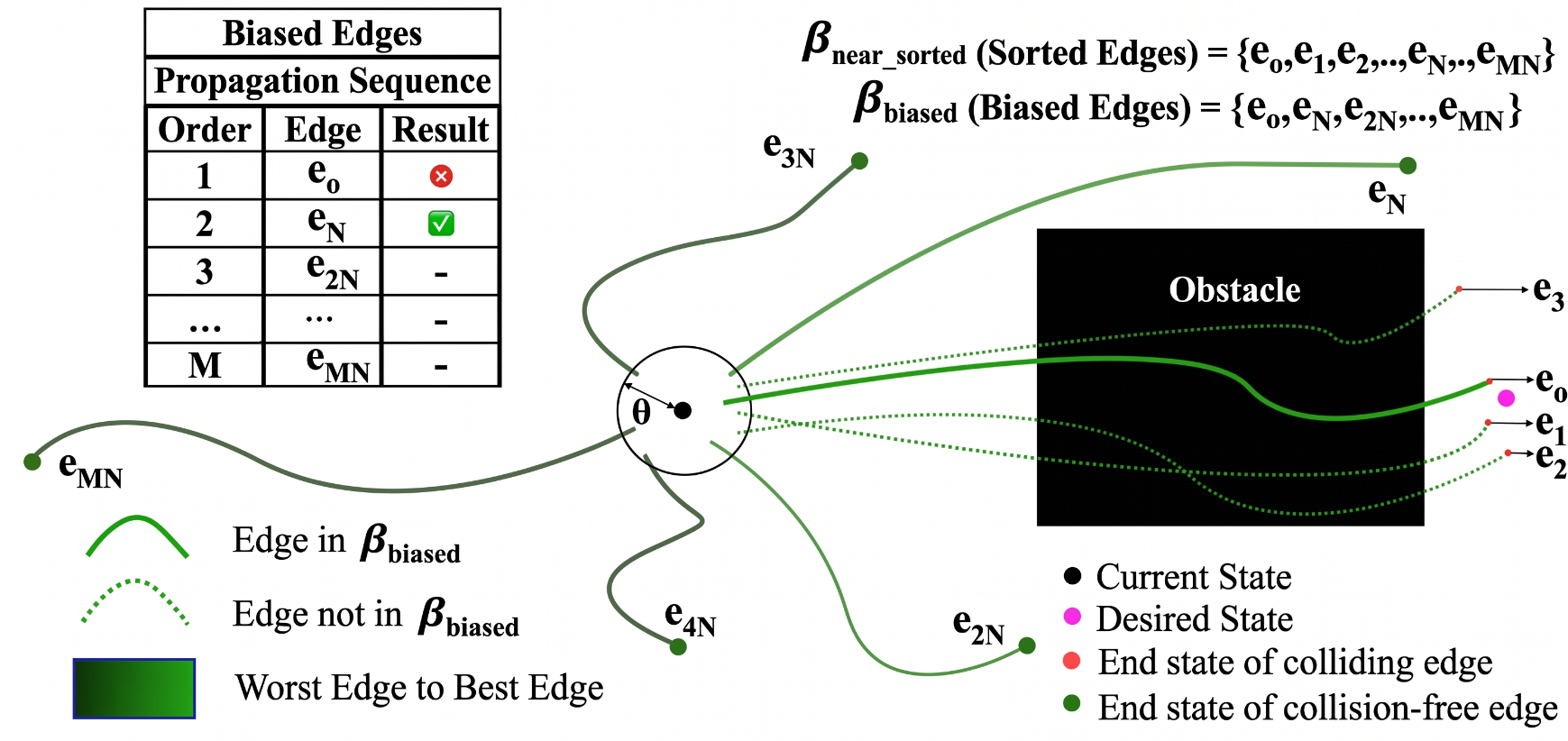}
}
\caption{Given the robot's current state, we gather all edges initiating within a radius of $\theta$ from it and store them in $\boldsymbol{\beta}_{\text{\upshape near\_sorted}}$. From this set, we extract relevant edges and store them in sets $\boldsymbol{\beta}_{\text{\upshape random}}$ (a) and $\boldsymbol{\beta}_{\text{\upshape biased}}$ (b). We show the exact order of edge propagation in each scenario with the result.}
\label{fig:strategy}
\end{figure}
To expand a tree node $x_\text{\upshape near}$ toward a desired state $x_\text{\upshape des}$, the planner considers the neighboring edges (i.e., those whose start state $x_0$ lies within a radius $\theta$ of $x_\text{\upshape near}$). After a rigid-body transformation of each such edge's pose components onto $x_\text{\upshape near}$, the stored end state $x_f$ serves only to \textsl{rank} the candidates. The stored control $u$ and duration $\Delta t$ are then propagated from the actual node $x_\text{\upshape near}$ (i.e., the true dynamics are integrated online). Consequently, a mismatch between $x_0$ and $x_\text{\upshape near}$, which $\theta$ keeps bounded, degrades only the ranking heuristic and never the feasibility of the added trajectory. This is because every tree edge is exact and free of discontinuities. Because the pose components are transformed away, the bundle only needs to cover the kinodynamic subspace (velocities, steering and hitch angles) rather than the full state space. However, adding obstacles to the state space after the edge generation phase results in several edges colliding with these obstacles during the planning phase. Next, we try to overcome this issue with a strategic approach.
\subsection{Edge Selection Strategy}
\label{sec:strategy}
An expansive and dense edge bundle can guide the robot from its current state to the desired state.
For this purpose, BoE \cite{BoE} propagates all the edges in the vicinity of the current state, resulting in excessive computation when new obstacles are introduced.
 As seen in \cref{fig:strategy} (b), once we gather all the edges that initiate within a neighborhood of radius $\theta$ from the robot's current state, we don't have to spend redundant computation by propagating all of them while aiming to reach the desired state. A more computationally efficient approach is to sort them based on the distance between their end state and the desired state, resulting in $\boldsymbol{\beta}_{\text{\upshape near\_sorted}}$ $(\{\mathrm{e}_0, \mathrm{e}_1, \mathrm{e}_2, \ldots \})$. They can then be propagated one by one until a collision-free trajectory is obtained. However, if the 
best edge $\text{e}_\text{0}$, that gets the closest to the desired state, collides with an obstacle that was not present during edge bundle generation, the next best edges $\text{e}_\text{1}$, $\text{e}_\text{2}$ and $\text{e}_\text{3}$ have a high chance of colliding with the same obstacle, due to the extensive number of edges.\\
Our strategy maximizes the reach towards the desired state while minimizing the number of propagation attempts.
To do so, we choose a user-defined integer $\text{N} \geq 1$ and let $\text{M} = \lfloor (|\boldsymbol{\beta}_{\text{\upshape near\_sorted}}|-1)/\text{N} \rfloor$, so that $\text{MN}$ is the largest multiple of $\text{N}$ that indexes a sorted edge. We pick the best edge (i.e., $\text{e}_\text{0}$) and every $\text{N}$th edge from the beginning of $\boldsymbol{\beta}_{\text{\upshape near\_sorted}} $(i.e., $\text{e}_\text{N},\text{e}_\text{2N}, \text{e}_\text{3N}, \ldots,\text{e}_\text{MN}$) and add them to a list $\boldsymbol{\beta}_{\text{\upshape biased}} $, giving rise to $\text{M+1}$ biased edges ($\{\mathrm{e}_\text{\upshape 0},\mathrm{e}_\text{\upshape N}, \mathrm{e}_\text{\upshape 2N}, \mathrm{e}_\text{\upshape 3N}, \ldots , \mathrm{e}_\text{\upshape MN} \}$).
 Essentially, if the best edge $\text{e}_\text{0}$ undergoes a collision, we are skipping $N-1$ edges that have a high chance of colliding with the same obstacle, thus avoiding redundant computation. Until a collision-free trajectory is obtained, we attempt to propagate every Nth edge until we reach the last biased edge (i.e., $\text{e}_\text{MN}$), which is the worst-ranked edge we consider. In heavily cluttered spaces, there is a high probability that most of the edges in $\boldsymbol{\beta}_{\text{\upshape biased}} $ lead to collisions, resulting in very few successful expansions. The trajectories that are successful could either traverse away from the desired state or converge very little towards it. This is problematic whenever the current and desired states remain fixed, as in the exploitation phase of \cref{fig:architecture}.\\
 To tackle this issue, we refer to Fig. \ref{fig:strategy} (a), where $\text{M+1}$ edges (i.e., $\text{e}^1_\text{r},\text{e}^2_\text{r},\ldots,\text{e}^\text{M+1}_\text{r}$), as many as in $\boldsymbol{\beta}_{\text{\upshape biased}}$, are drawn uniformly without replacement from $\boldsymbol{\beta}_{\text{\upshape near\_sorted}}$ and stored in $\boldsymbol{\beta}_{\text{\upshape random}}$.
 This is a better alternative in cluttered spaces for a fixed pair of current and desired states. Due to the inherent stochasticity, using the randomized edges (i.e., $\boldsymbol{\beta}_{\text{\upshape random}} $) might find feasible narrow paths absent in $\boldsymbol{\beta}_{\text{\upshape biased}}$ that eventually converge towards the goal. This strategy of choosing either $\boldsymbol{\beta}_{\text{\upshape biased}}$ or 
 $\boldsymbol{\beta}_{\text{\upshape random}} $ helps guide the robot to efficiently converge towards the goal amidst clutter. Next, we discuss our planner, BBoE, which incorporates this strategy.
\subsection{ Bidirectional Bundle of Edges}
\setlength{\textfloatsep}{2pt plus 1pt minus 1pt}
\setlength{\floatsep}{2pt plus 1pt minus 1pt}
\setlength{\intextsep}{2pt plus 1pt minus 1pt}
\RestyleAlgo{ruled}
\SetKwComment{Comment}{/* }{ */}
\begin{algorithm}[t] \scriptsize
\caption{ BBoE$(x_{\text{start}}, x_{\text{goal}},\delta_{\text{hr}}, k_\text{\upshape bias}, k_\text{\upshape exploit}, \text{\upshape N})$}\label{alg:one}
$\it{G_{\text{for}}} \leftarrow \{\textbf{V}_{\text{for}} \leftarrow \{x_{\text{start}}\}, \textbf{E}_{\text{ for}} \leftarrow \{\} \}$\\
$\it{G_{\text{rev}}} \leftarrow \{\textbf{V}_{\text{rev}} \leftarrow \{x_{\text{goal}}\}, \textbf{E}_{\text{ rev}} \leftarrow \{\} \}$\\
$\textbf{Q} \leftarrow \{\}$, $k_{\text{exploration}} \leftarrow 1$\\
\textcolor{Black}{$\boldsymbol{\beta} \leftarrow$ LoadEdgeBundle()}\\
 \For{$k \leftarrow 1 $ \KwTo $M_{iter}$}{
    $r_k \leftarrow$ min$(\gamma(\text{log}(|\textbf{V}_{\text{rev}}|)/|\textbf{V}_{\text{rev}}|)^{1/{d+1}}, \delta_{\text{hr}})$\\
$x_{\text{rand}} \leftarrow \text{RandomState()}$\\
$x_{\text{near}} \leftarrow \text{NearestNeighbour}(x_{\text{rand}}, \it{G_{\text{rev}}} )$\\
\textcolor{Black}{$\varepsilon_{\text{rev}} \leftarrow \text{RevPropUsingEdgeBundle}(\boldsymbol{\beta}, x_{\text{near}}, x_{\text{rand}}, \text{\upshape N})$}\\
\If{\textcolor{Black}{$\varepsilon_{\text{rev}} \neq  \text{ \upshape NULL}$}}{
    $x_{\text{rev}} \leftarrow \varepsilon_{\text{rev}}.\text{finalState}()$\\
    $\textbf{E}_{\text{ rev}} \leftarrow \textbf{E}_{\text{ rev}} \cup \{\varepsilon_{\text{rev}}\}$\\
    $\textbf{V}_{\text{rev}} \leftarrow \textbf{V}_{\text{rev}} \cup \{x_{\text{rev}}\}$\\
    $\text{updatePriorityQueue}(\it{G_{\text{for}}}, \textbf{Q}, x_{\text{rev}}, r_k)$
    \
    }
$q \leftarrow k_\text{\upshape exploit}$\\
$c_{\text{rand}} \sim  U([0,1])$\\
\If{$c_{\text{rand}} < q$}{
    $\varepsilon_{\text{for}} = \text{NULL}$\\
    $x_{\text{pop}} \leftarrow \text{Pop}(\textbf{Q})$\\
    \If{$x_{\text{pop}} \neq \text{\upshape NULL}$}{
        $\textbf{V}_{\text{rev\_near}} \leftarrow \{x \in \it{G_{\text{rev}}}.\text{Nodes}()~ |~ d_{\text{X}}(x_{\text{pop},x}) \leq r_k  \}$\\
        $x_{\text{best}} \leftarrow \text{arg min}_{x\in \textbf{V}_{\text{rev\_near}}}(g(x_{\text{pop}})+d_{\text{X}}(x_{\text{pop},x}) + h(x))$\\
        \textcolor{Black}{$\varepsilon_{\text{for}} \leftarrow \text{BestPropUsingEdgeBundle}(\boldsymbol{\beta}, x_{\text{pop}}, x_{\text{best}}, k_\text{\upshape bias}, \text{\upshape N})$} 
        }
    \If{$\varepsilon_{\text{for}} = \text{\upshape NULL}$}{
        $x_{\text{rand}} \leftarrow \text{RandomState()}$\\
        $x_{\text{near}} \leftarrow \text{NearestNeighbour}(x_{\text{rand}}, \it{G_{\text{for}}} )$\\
        \textcolor{Black}{$\varepsilon_{\text{for}} \leftarrow \text{BestPropUsingEdgeBundle}(\boldsymbol{\beta}, x_{\text{near}}, x_{\text{rand}}, k_{\text{exploration}}, \text{\upshape N})$}\\
        } 
     }
    
\If{$c_{\text{rand}} \geq q$  \textbf{\upshape or} $\varepsilon_{\text{for}} = \text{\upshape NULL}$}{
        $x_{\text{rand}} \leftarrow \text{RandomState()}$\\
        $x_{\text{near}} \leftarrow \text{NearestNeighbour}(x_{\text{rand}}, \it{G_{\text{for}}} )$\\
        \textcolor{Black}{$\varepsilon_{\text{for}} \leftarrow \text{RandomPropUsingEdgeBundle}(\boldsymbol{\beta}, x_{\text{near}})$}
        }
\If{$\varepsilon_{\text{for}} \neq \text{\upshape NULL}$}{
        $x_{\text{for}} \leftarrow \varepsilon_{\text{for}}.\text{finalNode}()$\\
        $\textbf{E}_{\text{ for}} \leftarrow \textbf{E}_{\text{ for}} \cup \{\varepsilon_{\text{for}}\}$\\
        $\textbf{V}_{\text{for}} \leftarrow \textbf{V}_{\text{for}} \cup \{x_{\text{for}}\}$    \\
        \If{$\text{\upshape goalRegionReached}(x_{\text{goal}},x_{\text{for}})$}{
            $\textbf{return} ~ \text{Path}(\it{G_{\text{for}}},x_{\text{start}}, x_{\text{for}}) $
            \ }
        $\text{InsertToPriorityQueue}(\it{G_{\text{rev}}}, \textbf{Q}, x_{\text{for}}, r_k)$
        \ }
}
\textbf{return} \text{NULL}

\end{algorithm}

\subsubsection{Inputs (Alg. 1)}
\textsl{BBoE} takes the robot's start state $x_\text{start}$, any goal state within the goal region ($x_\text{goal} \in Q_{goal}$), and four user-defined parameters, $\delta_{\text{hr}}$, $k_\text{\upshape bias}$, $k_\text{\upshape exploit}$ and $\text{\upshape N}$ as input. $\delta_{\text{hr}}$ is the maximum distance within which nodes from the reverse tree have a focusing effect on the growth of the forward tree, $k_\text{\upshape bias}$ is the probabilistic threshold used to decide between $\boldsymbol{\beta}_{\text{\upshape biased}}$ and $\boldsymbol{\beta}_{\text{\upshape random}}$, and $\text{\upshape N}$ is the interval for choosing edges for $\boldsymbol{\beta}_{\text{\upshape biased}}$. $k_\text{\upshape exploit}$ is the constant probability with which an iteration attempts exploitation (Line 15). 
\subsubsection{Initialization (Alg. 1, Lines 1–4)}
We use the node set $\textbf{V}_{\text{for}}$ (initialized to $x_{\text{start}}$) and the edge set $\textbf{E}_{\text{for}}$ to initialize the forward search tree $\it{G_{\text{for}}}$ (line 1). Similarly, we use the node set $\textbf{V}_{\text{rev}}$ (initialized to $x_{\text{goal}}$ ) and the edge set $\textbf{E}_{\text{rev}}$ to initialize the reverse tree $\it{G_{\text{rev}}}$ (line 2). We create a priority queue $\textbf{Q}$ to maintain a list of prospective nodes in the forward tree to use for tree expansion. We also initialize $k_\text{exploration}$ to 1, which ensures that BestPropUsingEdgeBundle only uses $\boldsymbol{\beta}_{\text{\upshape biased}}$ during exploration (line 3). We conclude the initialization segment by loading a precomputed bundle of edges $\boldsymbol{\beta}$ (line 4).
\subsubsection{Shrinking Neighborhood Radius $r_k$ (Alg. 1, Line 6)}
Radius $r_k$ has two objectives. It is used to determine if the trees are close enough to shift from exploration to exploitation (line 21) and to
influence the priority queue update and insertion operations (lines 14 and 38). Here, $d$ is the dimension of the state space and $\gamma$ is a scenario-specific parameter. 
\subsubsection{Reverse Search (Alg. 1, Lines 7-14)}
We use the edge bundle $\boldsymbol{\beta}$ for expanding the reverse tree. The aim is to quickly explore the state space and provide a heuristic to the forward tree for goal convergence. First, we randomly sample a state (Line 7) and identify the nearest node in the reverse tree (Line 8). Then, RevPropUsingEdgeBundle propagates the nearest node towards the random state using the bundle of edges (Line 9). If the returned edge $\varepsilon_{\text{rev}}$ is not NULL (Line 10), then we add $\varepsilon_{\text{rev}}$ and its corresponding final state $x_{\text{rev}}$ (Line 11) to the corresponding sets (Lines 12 and 13). We utilize $x_{\text{rev}}$ to perform a heuristic update in $\textbf{Q}$ using updatePriorityQueue (Line 14).
\subsubsection{Forward Search (Alg. 1, Lines 15-39)}
The forward search is an active part of both the exploration phase (Lines 24-27 and 28-31) and exploitation phase (Lines 20-23). The exploitation probability $q$ is the constant $k_\text{\upshape exploit}$ (line 15).
The value of q decides if the exploration phase or the exploitation phase is triggered (Line 17). If the exploitation phase is initiated, we check if the priority queue is empty (Line 19). If so, then exploitation isn't possible and we perform exploration instead (Lines 24-27) by calling BestPropUsingEdgeBundle (Alg. 2), where we leverage the strategy on the edge bundle $\boldsymbol{\beta}$ to expand the tree towards the desired state. If exploitation wasn't chosen in Line 17, we perform exploration in a random direction (Lines 29-31) via RandomPropUsingEdgeBundle (Alg. 3). If the priority queue isn't empty and exploitation was preferred (Line 20), we begin the exploitation process by gathering all the reverse tree nodes that are within radius $r_k$ from $x_{pop}$ and storing them in $\textbf{V}_{\text{rev\_near}}$ (Line 21). Then, we pick the best node amongst $\textbf{V}_{\text{rev\_near}}$(Line 22) and the edge bundle $\boldsymbol{\beta}$ aids the propagation from $x_{pop}$
 to $x_{best}$ (Line 23). Once we obtain the new edge-node pair (Line 33), we add them to the tree (Lines 34 and 35) and check if we have reached the goal (Lines 36 and 37). If not, we add the new forward node to the priority queue $\textbf{Q}$ if it is within $r_k$ distance from any of the reverse tree nodes (Line 38).

\begin{algorithm}[t] \scriptsize
\caption{\footnotesize{\textcolor{Black}{BestPropUsingEdgeBundle$(\boldsymbol{\beta}, x_{\text{near}}, x_{\text{des}}, k_\text{bias}, \text{\upshape N})$}}}\label{alg:two}
\textcolor{Black}{$\boldsymbol{\beta}_{\text{near}} \leftarrow \{ \text{e} \in \boldsymbol{\beta} ~ | ~ \text{d}(x_0, x_{\text{near}}) \leq \theta,~ \text{e untried at } x_{\text{near}} \}$\\
$\boldsymbol{\beta}_{\text{near\_sorted}} \leftarrow \boldsymbol{\beta}_{\text{near}}.\text{SortEdges}(x_{\text{des}}) $\\
$\boldsymbol{\beta}_{\text{biased}} \leftarrow \{ e_{\text{i}} \in \boldsymbol{\beta}_{\text{near\_sorted}} ~ | ~ i = 0,\text{\upshape N},\text{\upshape 2N},\ldots,\text{\upshape MN} \}$
\\
$\boldsymbol{\beta}_{\text{random}} \leftarrow \{ e_{\text{i}} \in \boldsymbol{\beta}_{\text{near\_sorted}} ~ | ~ i \sim U(\{0,\ldots,|\boldsymbol{\beta}_{\text{near\_sorted}}|-1\}) \}$, $|\boldsymbol{\beta}_{\text{random}}| = \text{M+1}$, no repeats\\
$q \leftarrow k_\text{bias}$\\
$c_{\text{rand}} \sim U([0,1])$\\
\If{$c_{\text{rand}} \leq q$}{
        $\boldsymbol{\beta}_{\text{final}} \leftarrow  \boldsymbol{\beta}_{\text{biased}}$
        \ }
        \Else{
            $\boldsymbol{\beta}_{\text{final}} \leftarrow  \boldsymbol{\beta}_{\text{random}}$
            } 
\While{$\boldsymbol{\beta_{\text{final}} \neq \phi}$}
    {
        $e \leftarrow \text{Pop}(\boldsymbol{\beta}_{\text{final}})$\\
        $\varepsilon_{\text{new}} \leftarrow  e.\text{Propagate}(x_{\text{near}})$\\
        \If{$\varepsilon_{\text{\upshape new}}.\text{\upshape Success()}$}
            {\textbf{return} $\varepsilon_{\text{\upshape new}}$  
            \ }
    }
\textbf{return} NULL
}
\end{algorithm}
\begin{algorithm}[t] \scriptsize
\caption{\textcolor{Black}{RandomPropUsingEdgeBundle$(\boldsymbol{\beta}, x_{\text{near}})$}}\label{alg:three}
\textcolor{Black}{$\boldsymbol{\epsilon}_{\text{near}} \leftarrow \{ \text{e} \in \boldsymbol{\beta} ~ | ~ \text{d}(x_0, x_{\text{near}}) \leq \theta,~ \text{e untried at } x_{\text{near}} \}$\\
$e \sim U(\boldsymbol{\epsilon}_{\text{near}})$\\
 $\varepsilon_{\text{new}} \leftarrow  e.\text{Propagate}(x_{\text{near}})$\\
        \If{$\varepsilon_{\text{\upshape new}}.\text{\upshape Success()}$}
            {\textbf{return} $\varepsilon_{\text{\upshape new}}$  
            \ }
\textbf{return} NULL
}
\end{algorithm}
\subsubsection{Propagation using Edge Bundles (Alg. 2)}
\label{sec:propagate}
Propagate (Alg.~2, Line 13) integrates the edge's stored control for its stored duration from the tree node $x_\text{\upshape near}$ using the true dynamics $f$, collision-checks the resulting trajectory, and returns it with a success flag. For the reverse tree, Propagate integrates backward in time (the duration is negated), so that a reverse node is a state from which its parent is reached under the stored control. The cost stored on a reverse node is the forward-time cost to the goal. Because the stored end state $x_f$ describes forward propagation, it does not rank edges for backward propagation. The reverse tree therefore calls RevPropUsingEdgeBundle$(\boldsymbol{\beta}, x_\text{\upshape near}, x_\text{\upshape rand}, \text{N})$ (Alg.~1, Line 9), which neither sorts nor builds $\boldsymbol{\beta}_{\text{\upshape biased}}$. It draws $\lfloor |\boldsymbol{\beta}_{\text{\upshape near}}^{\text{\upshape untried}}|/\text{N} \rfloor$ (at least one) of the node's untried neighboring edges at random for $\text{N} \geq 2$, or $\lfloor |\boldsymbol{\beta}_{\text{\upshape near}}|/5 \rfloor$ of them when $\text{N}=1$, where 5 is a fixed constant. Here, it propagates each edge backward and keeps the one whose end state is closest to $x_\text{\upshape rand}$ under $d_X$. These propagations per reverse expansion are included in all reported planning times. BestPropUsingEdgeBundle utilizes the edge bundle $\boldsymbol{\beta}$, user-defined values $k_\text{\upshape bias}$ and N to propagate from the current state ($x_\text{\upshape near}$) to the desired state ($x_\text{\upshape des}$).
$k_\text{\upshape bias}$ is a number between 0 and 1 providing a probabilistic threshold for the planner to choose $\boldsymbol{\beta}_{\text{biased}}$ over $\boldsymbol{\beta}_{\text{random}}$. During exploration, $k_\text{\upshape bias}$ will be hardcoded to 1 in order to impose the usage of $\boldsymbol{\beta}_{\text{biased}}$ and during exploitation, it will hold a user-defined value. N refers to the interval between sorted edges in $\boldsymbol{\beta}_{\text{\upshape biased}}$. We begin by focusing on $\boldsymbol{\beta}_{\text{\upshape near}}$, in which we store all the edges $\text{e}$ whose start states $x_0$ lie within radius $\theta$ from $x_\text{\upshape near}$ (Line 1). Then, we sort $\boldsymbol{\beta}_{\text{\upshape near}}$ based on the distance between each edge's end state and $x_\text{\upshape des}$, resulting in $\boldsymbol{\beta}_{\text{near\_sorted}}$ (Line 2). Afterwards, we retrieve the first and every Nth element in $\boldsymbol{\beta}_{\text{near\_sorted}}$ and store them in $\boldsymbol{\beta}_{\text{biased}}$ (Line 3). Furthermore, we draw $\text{M+1}$ edges (as many as in $\boldsymbol{\beta}_{\text{biased}}$) uniformly without replacement from $\boldsymbol{\beta}_{\text{near\_sorted}}$ and store them in $\boldsymbol{\beta}_{\text{random}}$ (Line 4). We set $q = k_\text{\upshape bias}$ (Line 5) and use it to determine if $\boldsymbol{\beta}_{\text{random}}$ or $\boldsymbol{\beta}_{\text{biased}}$ will be used for propagation (Lines 5-10). Until a successful propagation occurs or the chosen list $\boldsymbol{\beta}_{\text{final}}$ becomes empty, we pop edges from $\boldsymbol{\beta}_{\text{final}}$ and propagate them from $x_\text{\upshape near}$ (Lines 11-14), eventually returning the successful edge (Line 15). If none of the edges lead to a successful propagation, the result is NULL (Line 16). 
\subsubsection{Random Propagation using Edge Bundles (Alg. 3)}
RandomPropUsingEdgeBundle is used to quickly expand the forward tree by propagating a random neighboring edge. We store all the edges $\text{e}$ whose start states $x_0$ lie within radius $\theta$ from $x_\text{\upshape near}$ in $\boldsymbol{\epsilon}_{\text{\upshape near}}$ (Line 1). We randomly pick an edge from $\boldsymbol{\epsilon}_{\text{\upshape near}}$ (Line 2) and propagate it from $x_\text{\upshape near}$ (Line 3). If we obtain a collision-free trajectory $\varepsilon_{\text{\upshape new}}$, we return it (Line 4 and 5). Otherwise, the result is NULL (Line 6).

\section{EVALUATION}
\subsection{Experimental Setup}
\begin{figure*}[!t]
\centering
\includegraphics[width=\linewidth]{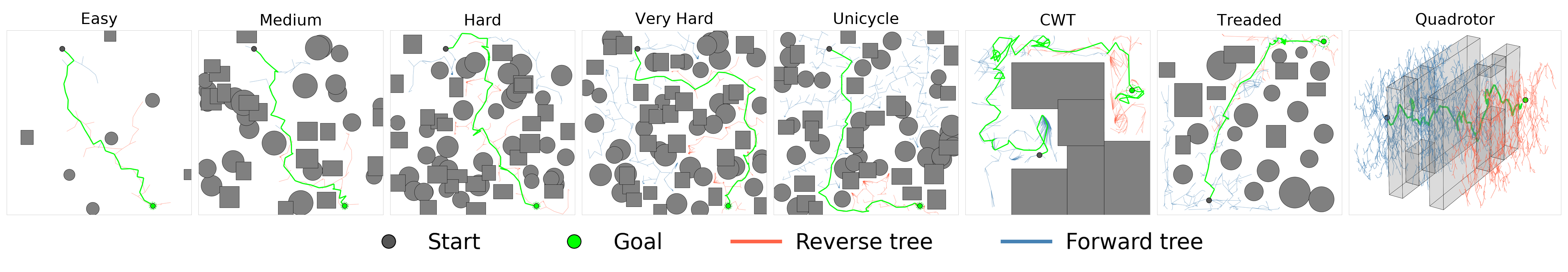}
\caption{Final BBoE solutions showing unicycle environments across four difficulty tiers (seed 15795) and four different dynamical systems. The forward tree (blue) traces the reverse tree (red) into the goal region. The solution path (green) consists of forward-tree edges.}
\label{fig:solutions}

\end{figure*}

\begin{figure}[hbt!]
\centering
\includegraphics[width=0.85\linewidth]{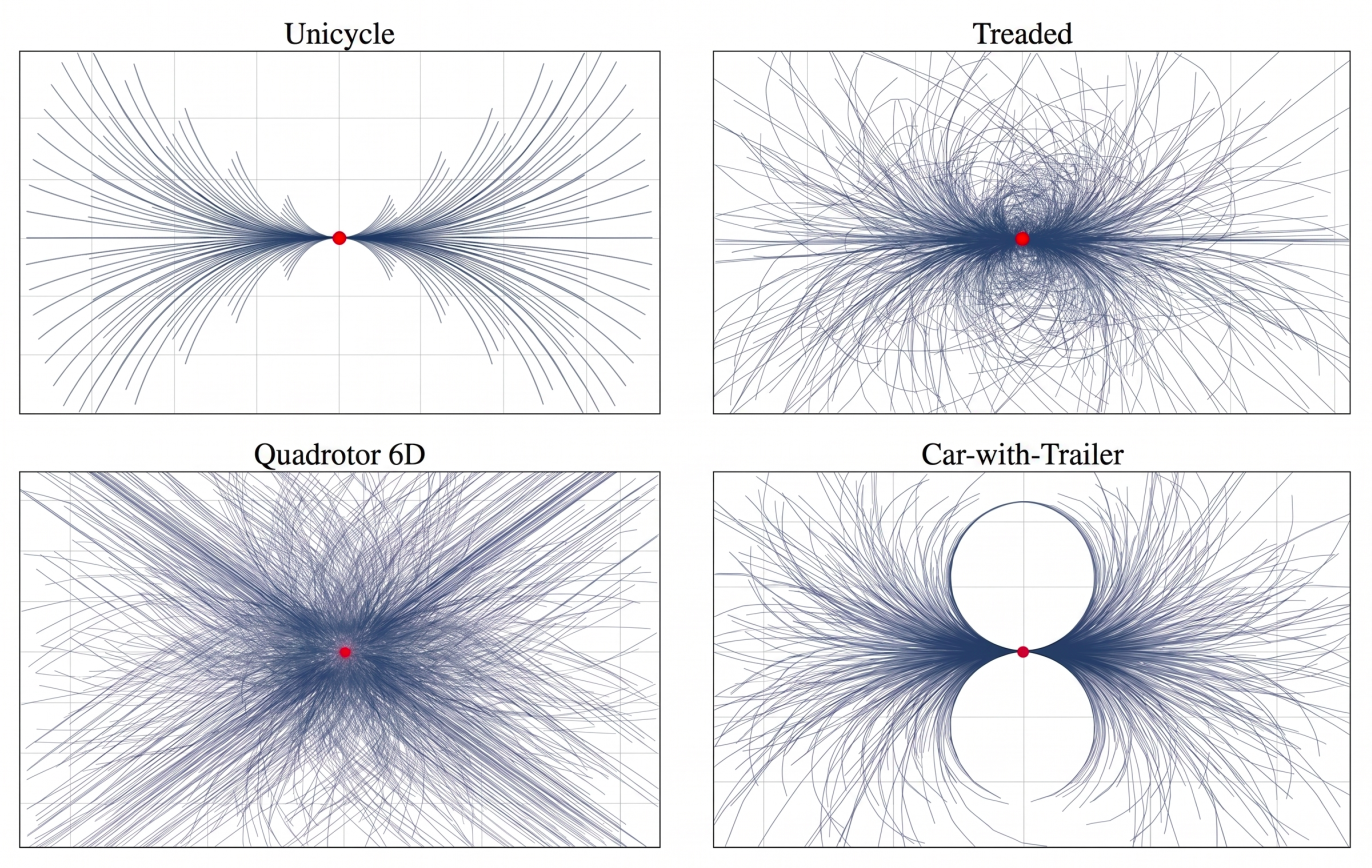}
\caption{{The 2-dimensional projection of a subset of the edges generated for each system. Every edge starts at the origin of the XY plane. The unicycle edges are essentially the maneuver libraries used in the GBRRT paper.}
}
\label{fig:edgebundles}

\end{figure}

As seen in Fig. \ref{fig:solutions}, we evaluate BBoE with $k_\text{\upshape bias} = 0.85$ on four systems of increasing complexity: a Unicycle, a Treaded Vehicle, a Car With Trailer (CWT), and a Quadrotor. The specifications of each system are shown in Table \ref{tab:system_specs}. The unicycle is kinematic with controls $(v,\omega)$. The treaded vehicle and the CWT are the second-order models of \cite{GBRRT}, the CWT state being $(x,y,\theta,v,\phi,\theta_1)$ (position, heading, speed, steering and trailer angle) with controls $(a,\dot{\phi})$. The quadrotor is a translational double integrator with attitude abstracted \cite{quad}, i.e., state $(\mathbf{p},\mathbf{v})$ and controls $\mathbf{a}$ with $|v_i|\leq0.5$, $|a_i|\leq2$. We compare BBoE against RRT \cite{RRT}, SST \cite{sst}, GBRRT \cite{GBRRT}, BoE \cite{BoE}, and Db-RRT \cite{idb_rrt}. We evaluate Db-RRT-F and Db-RRT-C without their subsequent trajectory optimization step. Db-RRT's trajectory-optimization step adds hundreds of milliseconds to several seconds~\cite{idb_rrt}. Since our focus is rapid initial-solution time and BBoE produces collision-free, dynamically feasible trajectories directly from search, we compare against the search component of Db-RRT in isolation. For the unicycle, we additionally evaluate Monte-Carlo (MC) and Edge-Bundle (EB) variants of RRT, SST, and GBRRT. The MC variants propagate online by generating 40 random control extensions per expansion and selecting the collision-free trajectory that terminates closest to the desired state. Likewise, the EB variants sample 40 edges from the precomputed bundle, propagate them from the current state, and select the expansion terminating closest to the desired state. Table \ref{tab:system_specs} lists the number of edges actually used at planning time.
\vspace{3pt}
\begin{table}[hbt!]
\centering
\footnotesize
\setlength{\tabcolsep}{4pt}
\renewcommand{\arraystretch}{1.25}
\caption{Specifications of the systems used in the experiments. For the distance functions, \(\mathbf{p}\) and \(\mathbf{v}\) denote the position and velocity vectors of the respective systems. \# Edges is the number of bundle edges used at planning time.}
\label{tab:system_specs}
\begin{tabular}{|l|c|c|c|c|}
\hline
\textbf{System} & \makecell{\textbf{State}\\\textbf{Dim.}} & \textbf{Distance Function} (\(d_X\)) & \makecell{\textbf{\#}\\\textbf{Edges}} & \textbf{Ref.} \\
\hline
Unicycle & 3 & \(\| \Delta \mathbf{p} \|\) & 190 & \cite{GBRRT} \\
\hline
Treaded & 5 & \(\big(\| \Delta \mathbf{p} \|^2 + 0.25^2 \| \Delta \mathbf{v} \|^2\big)^{1/2}\) & 50000 & \cite{GBRRT} \\
\hline
CWT & 6 & \(\big(\| \Delta \mathbf{p} \|^2 + 0.25^2 \Delta v^2\big)^{1/2}\) & 50000 & \cite{GBRRT} \\
\hline
Quadrotor & 6 & \(\big(\| \Delta \mathbf{p} \|^2 + \| \Delta \mathbf{v} \|^2 \big)^{1/2}\) & 100000 & \cite{quad} \\
\hline
\end{tabular}
\end{table}
For the unicycle, we further analyze planner performance across four difficulty levels (easy, medium, hard, and very hard) defined by obstacle count. Each level comprises 10 randomly generated environments evaluated across 50 seeds. For all systems, we generate a bundle of feasible edges, as seen in Fig.~\ref{fig:edgebundles}. Each edge is produced by sampling admissible control inputs and propagating the system dynamics over a random duration. For the unicycle, we use the same maneuver library as the GBRRT paper~\cite{GBRRT}, generated by discretely sampling actions and time durations, to ensure a fair comparison. For all other systems, edges are sampled from continuous action and duration ranges. During planning, edges are transformed according to the robot's pose before propagation. Crucially, this transformation is applied only to the positional and orientation components of the state (e.g., $x, y, z, \theta$), while the kinodynamic components (velocities, accelerations, angular rates, etc.) are preserved as-is, since rigid-body transformations are well-defined on the configuration manifold but do not extend consistently to the control space. To ensure fairness, BoE, Db-RRT, and BBoE leverage the same bundle of edges. All planners are evaluated over 50 random seeds using three metrics: i) Initial Solution Time, ii) Initial Solution Cost (Path Length), and iii) Success Rate. Each planner is assigned a fixed planning budget of 5 seconds, after which the trial is counted as a 
failure. On the treaded vehicle, BBoE and GBRRT both exclude reverse-tree nodes already reached by the forward tree (within 1 unit of $d_X$) from later exploitation targets to avoid retargeting nodes.

\subsection{Results}
\label{sec:results}
\begin{figure*}[!t]
\centering
\newcommand{\plt}[2]{\includegraphics[width=#1\linewidth, trim=0 0 0 23, clip]{#2}}
\begin{subfigure}{0.3\textwidth}
  \centering\plt{1}{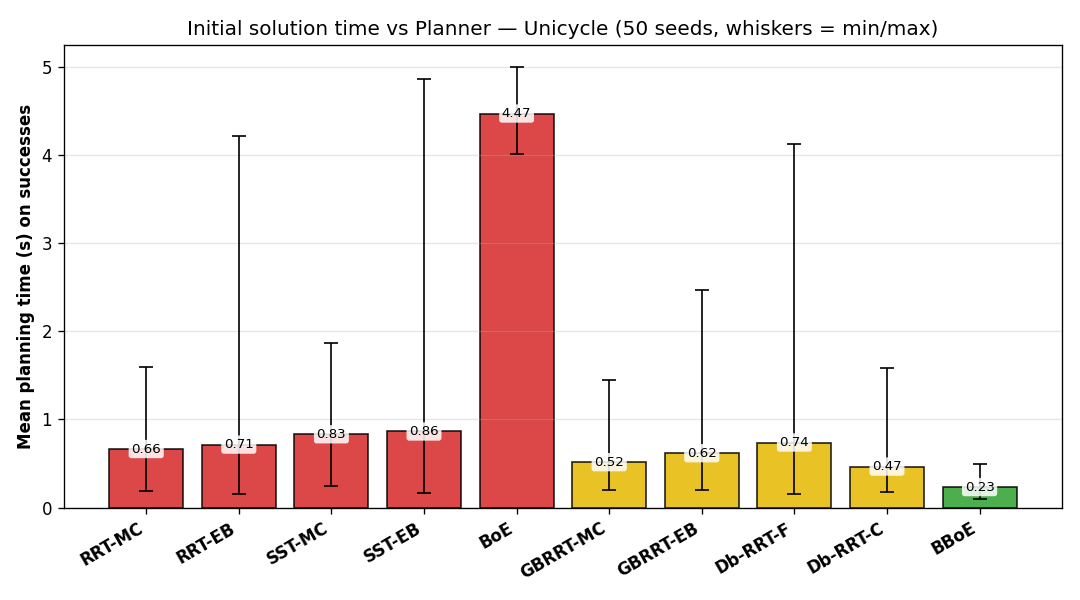}
  \caption{Unicycle: Solution Time}
\end{subfigure}\hfill
\begin{subfigure}{0.21\textwidth}
  \centering\plt{1}{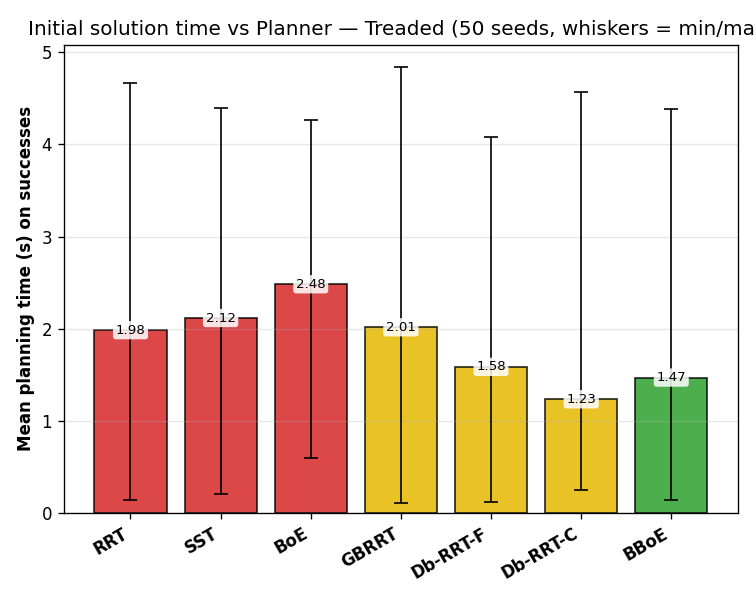}
  \caption{Treaded: Solution Time}
\end{subfigure}\hfill
\begin{subfigure}{0.21\textwidth}
  \centering\plt{1}{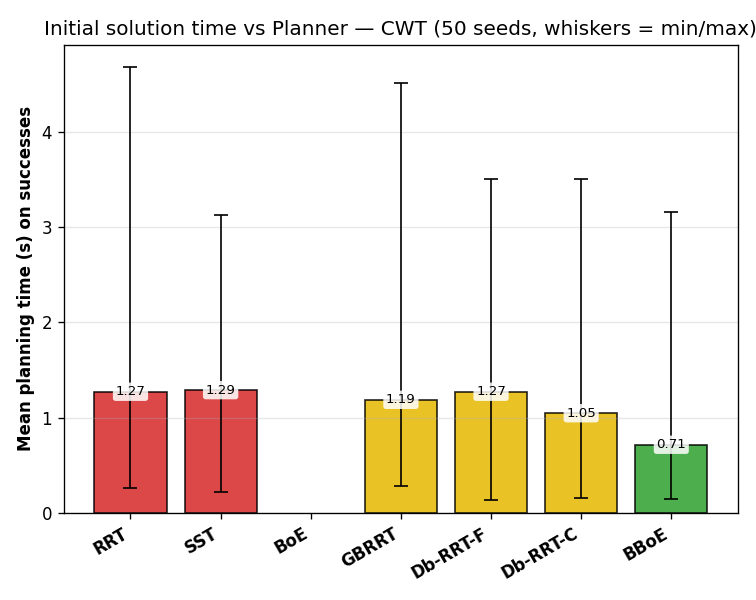}
  \caption{CWT: Solution Time}
\end{subfigure}\hfill
\begin{subfigure}{0.21\textwidth}
  \centering\includegraphics[width=\linewidth, trim=0 0 0 23, clip]{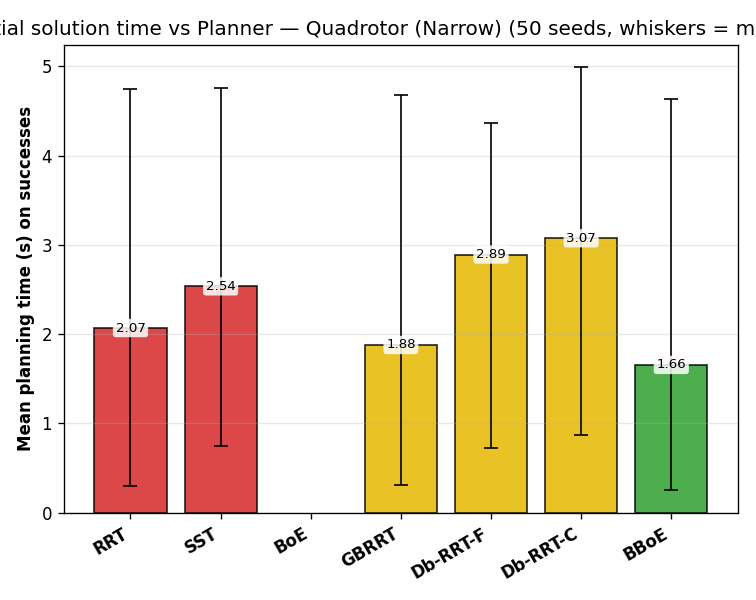}
  \caption{Quadrotor: Solution Time}
\end{subfigure}
\vspace{2pt}
\begin{subfigure}{0.3\textwidth}
  \centering\plt{1}{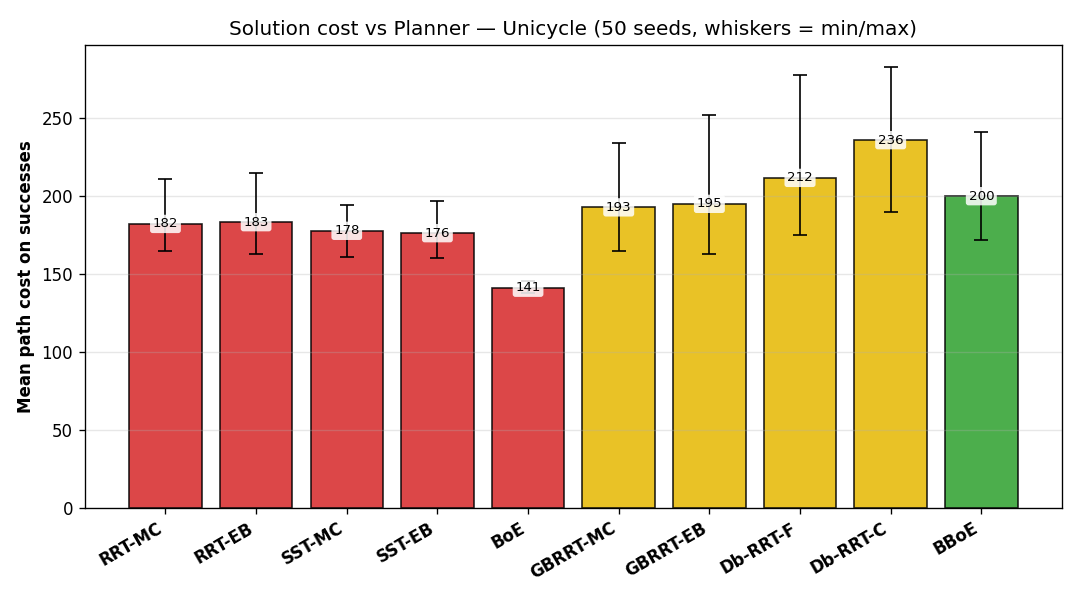}
  \caption{Unicycle: Solution cost}
\end{subfigure}\hfill
\begin{subfigure}{0.21\textwidth}
  \centering\plt{1}{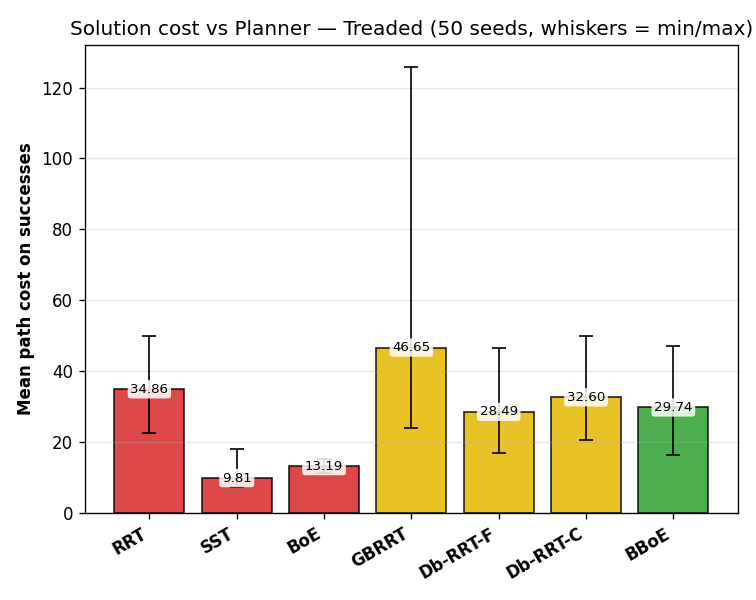}
  \caption{Treaded: Solution cost}
\end{subfigure}\hfill
\begin{subfigure}{0.21\textwidth}
  \centering\plt{1}{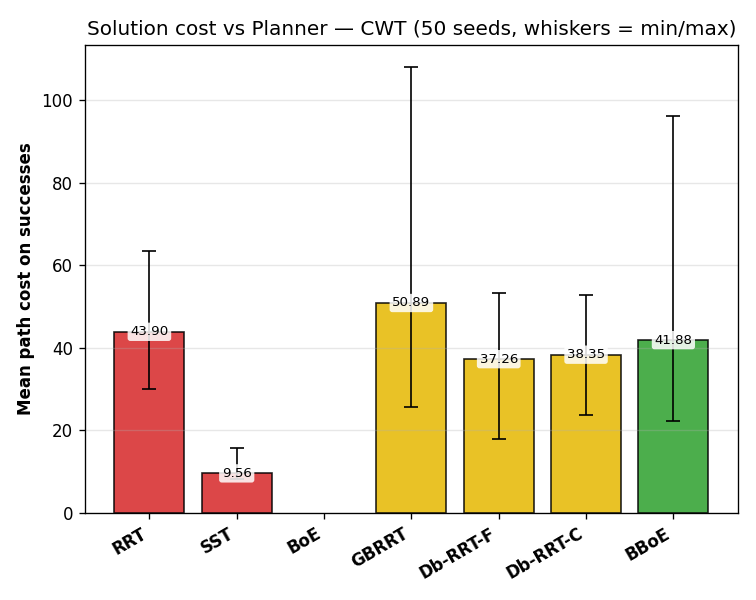}
  \caption{CWT: Solution cost}
\end{subfigure}\hfill
\begin{subfigure}{0.21\textwidth}
  \centering\plt{1}{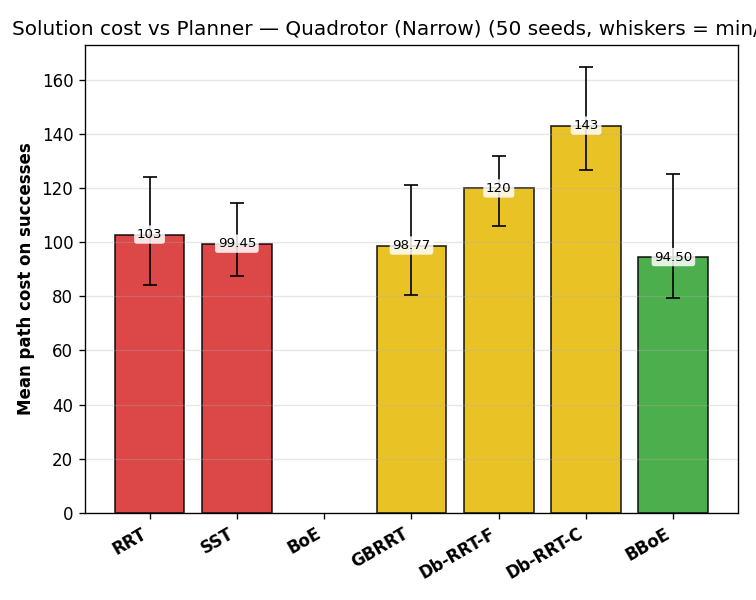}
  \caption{Quadrotor: Solution cost}
\end{subfigure}
\caption{Cross-system planner evaluation on the unicycle, treaded vehicle, car-with-trailer (CWT), and quadrotor (50 seeds each). \textbf{Top row:} mean initial solution time on successes per system. \textbf{Bottom row:} mean solution cost on successes per system. BBoE achieves the lowest planning time on three of four systems and $\geq96\%$ success on every system.}
\label{fig:cross_system_results}

\end{figure*}\begin{figure*}[!t]
\centering
\newcommand{\plt}[2]{\includegraphics[width=#1\linewidth, trim=0 0 0 23, clip]{#2}}
\begin{subfigure}{0.245\textwidth}
  \centering\plt{1}{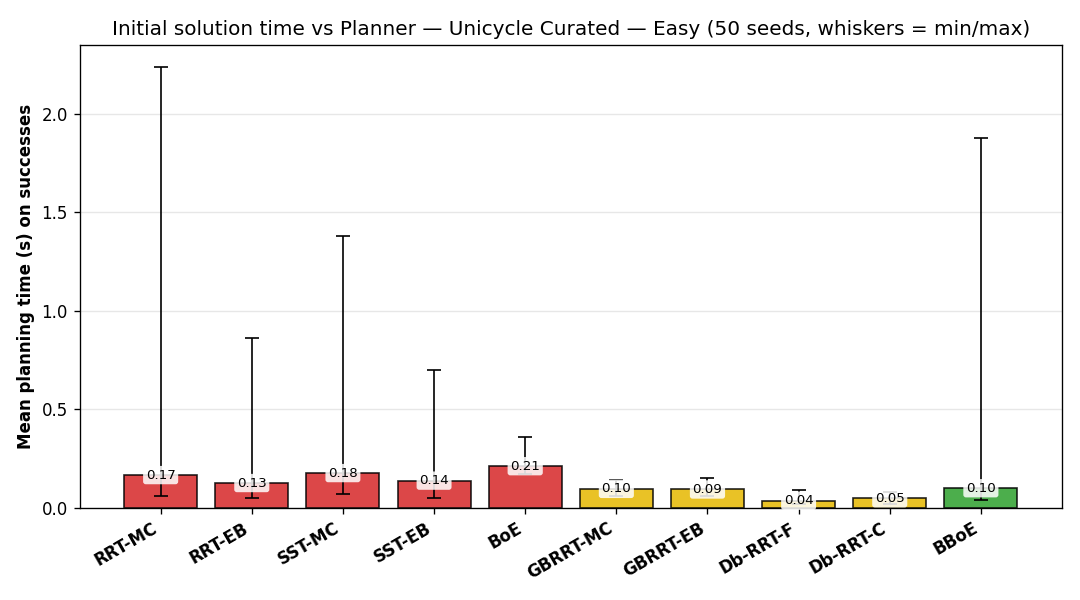}
  \caption{Easy: Solution Time}
\end{subfigure}\hfill
\begin{subfigure}{0.245\textwidth}
  \centering\plt{1}{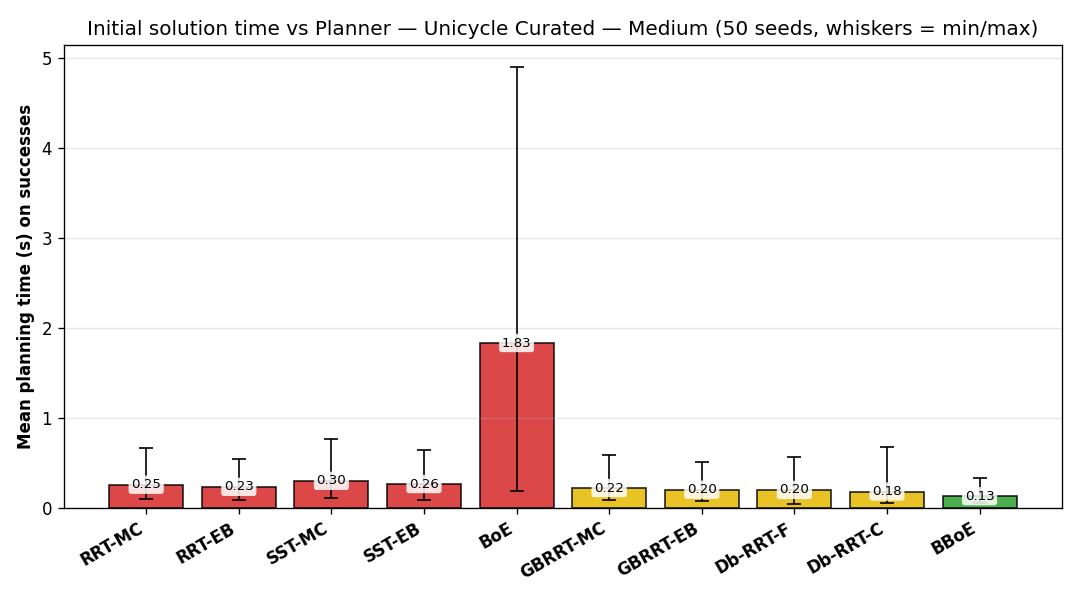}
  \caption{Medium: Solution Time}
\end{subfigure}\hfill
\begin{subfigure}{0.245\textwidth}
  \centering\plt{1}{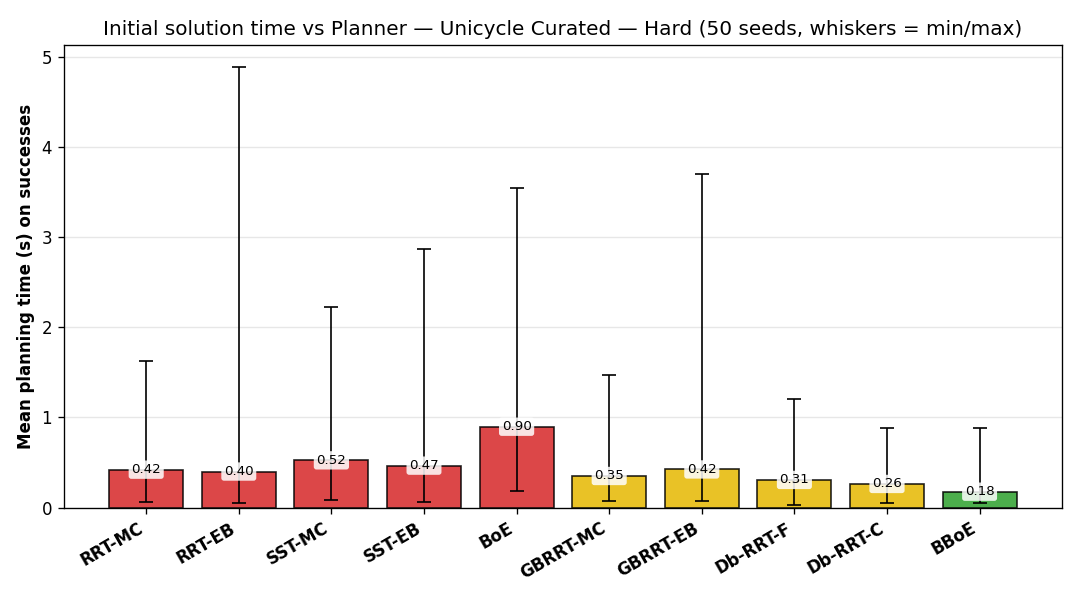}
  \caption{Hard: Solution Time}
\end{subfigure}\hfill
\begin{subfigure}{0.245\textwidth}
  \centering\plt{1}{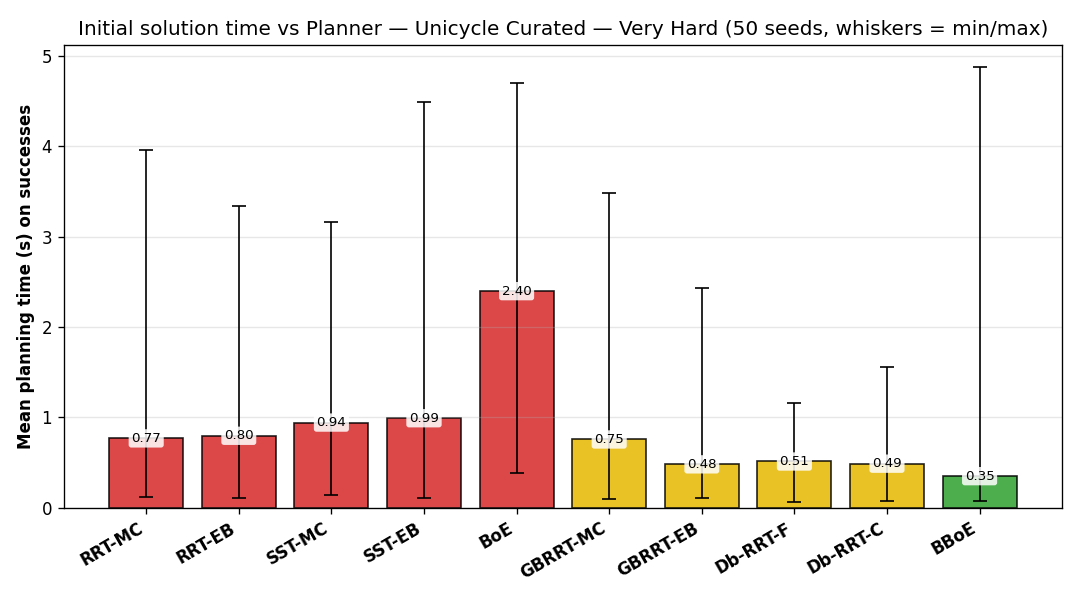}
  \caption{Very Hard: Solution Time}
\end{subfigure}
\vspace{2pt}
\begin{subfigure}{0.245\textwidth}
  \centering\plt{1}{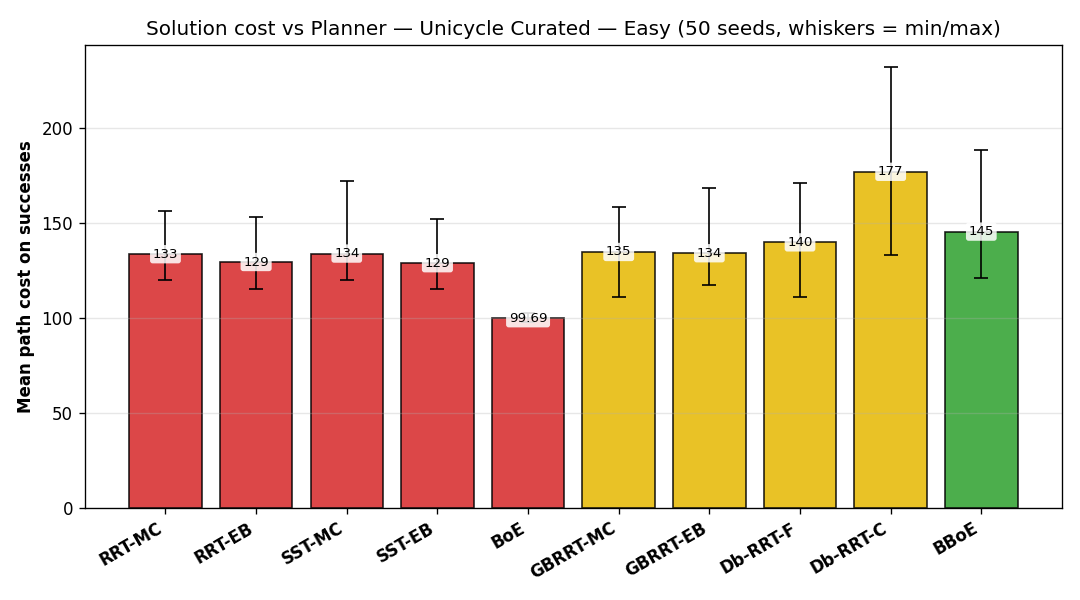}
  \caption{Easy: Solution Cost}
\end{subfigure}\hfill
\begin{subfigure}{0.245\textwidth}
  \centering\plt{1}{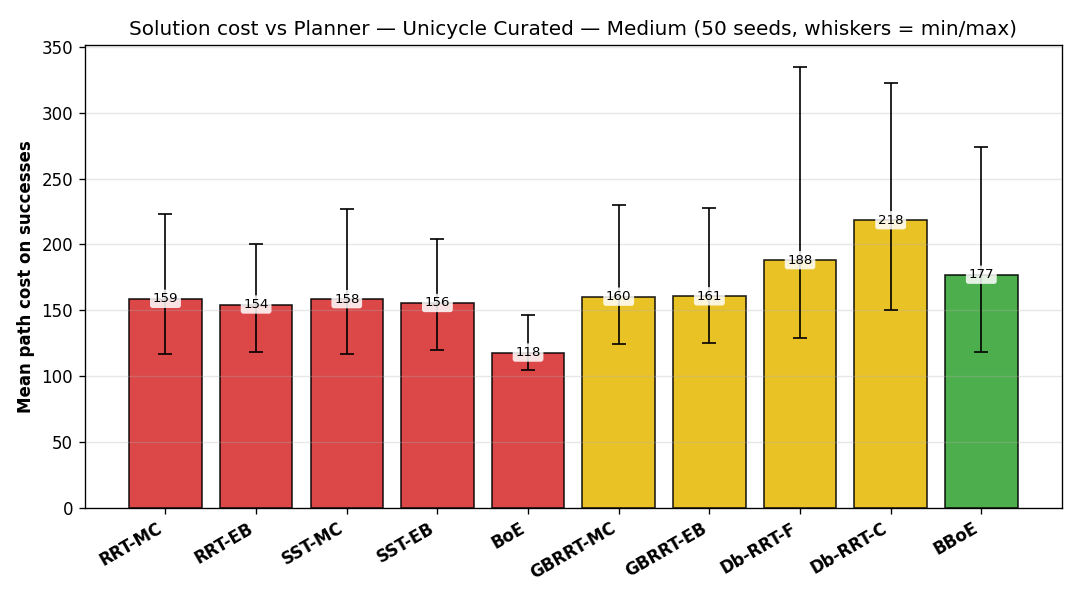}
  \caption{Medium: Solution Cost}
\end{subfigure}\hfill
\begin{subfigure}{0.245\textwidth}
  \centering\plt{1}{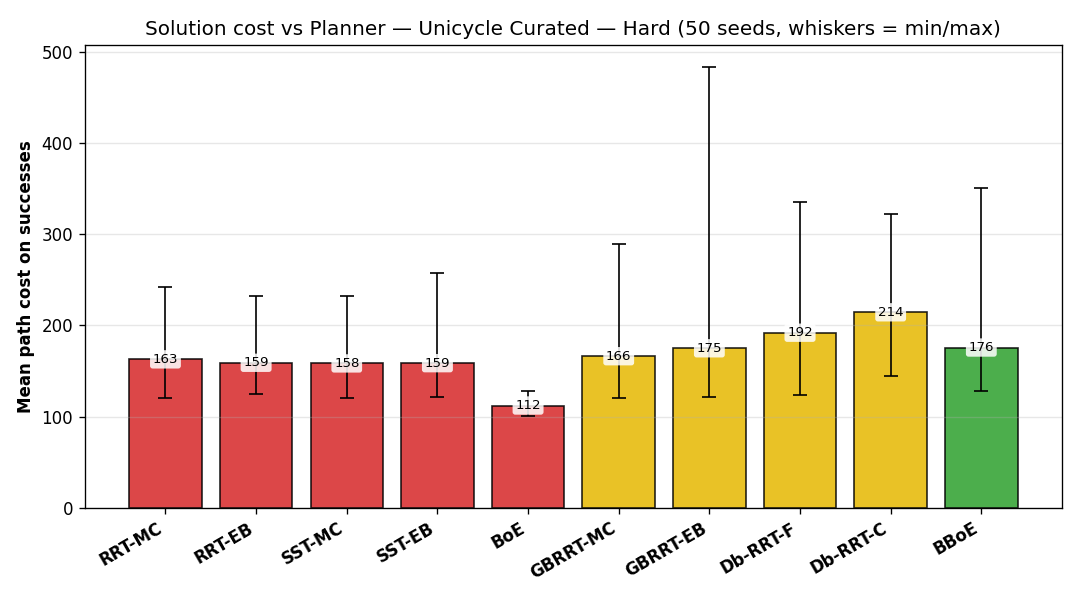}
  \caption{Hard: Solution Cost}
\end{subfigure}\hfill
\begin{subfigure}{0.245\textwidth}
  \centering\plt{1}{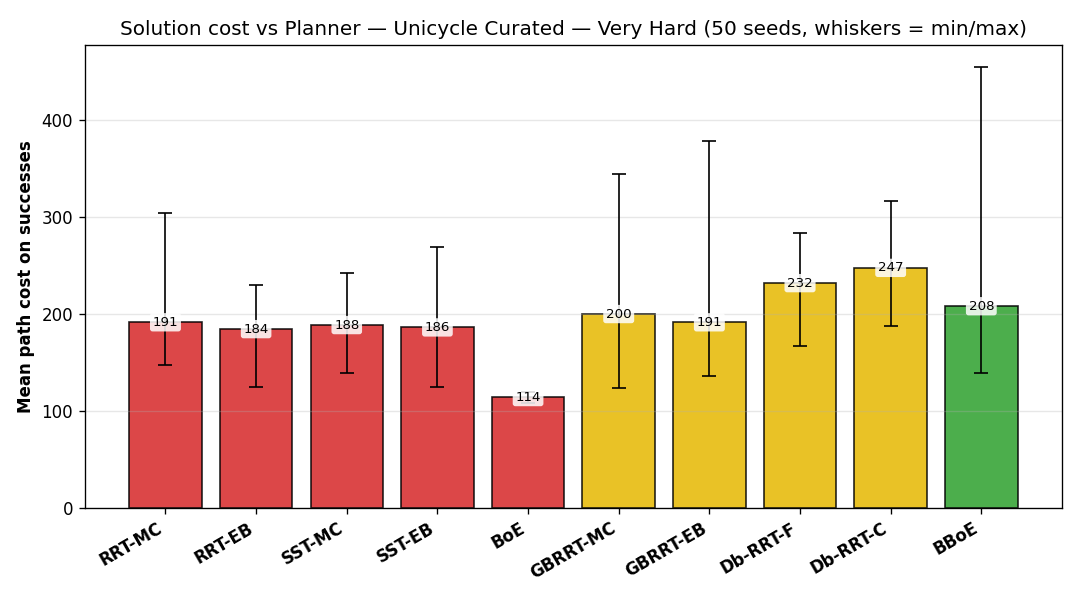}
  \caption{Very Hard: Solution Cost}
\end{subfigure}
\caption{Unicycle planner evaluation across 40 environments (10 per difficulty tier, 50 seeds each). \textbf{Top row:} Mean initial solution time on successes per difficulty. \textbf{Bottom row:} Mean solution cost on successes per difficulty.}
\label{fig:unicycle_curated_results}

\end{figure*}

Across all systems, BBoE consistently achieves the highest/ near-highest success rates and the lowest/ near-lowest planning times. BoE and Db-RRT, despite operating on the same edge bundles, do not achieve comparable speed in most settings. Substituting online Monte-Carlo propagation with offline edge reuse yields consistent improvements for RRT, SST, and GBRRT. However, the EB variants alone do not match BBoE's performance, suggesting that the observed gains stem from the interaction between edge reuse, bidirectional search, and BBoE's edge sequencing strategy rather than from edge reuse in isolation.
\begin{table}[hbt!]
\centering

\setlength{\tabcolsep}{2.5pt}
\renewcommand{\arraystretch}{1.0}
\tiny
\caption{Cumulative success rate (\%).}
\label{tab:success_rates}
\begin{tabular}{@{}l@{\hspace{3pt}}cccccccc@{}}
\toprule
Planner & E & M & H & VH & Uni & CWT & Tr & Qd \\
\midrule
RRT      & --  & --  & --  & --  & --  & 96  & 80  & 90  \\
RRT-MC   & 100 & 100 & 92  & 88  & 100 & --  & --  & --  \\
RRT-EB   & 100 & 100 & 100 & 90  & 100 & --  & --  & --  \\
SST      & --  & --  & --  & --  & --  & 100 & 82  & 52  \\
SST-MC   & 100 & 100 & 100 & 82  & 100 & --  & --  & --  \\
SST-EB   & 100 & 100 & 100 & 88  & 100 & --  & --  & --  \\
BoE      & 100 & 62  & 60  & 18  & 48  & 0   & 96  & 0   \\
GBRRT    & --  & --  & --  & --  & --  & 100 & 88  & 88  \\
GBRRT-MC & 100 & 100 & 98  & 100 & 100 & --  & --  & --  \\
GBRRT-EB & 100 & 100 & 100 & 100 & 100 & --  & --  & --  \\
Db-RRT-F & 100 & 100 & 90  & 52  & 100 & 86  & 78  & 24  \\
Db-RRT-C & 100 & 100 & 96  & 70  & 100 & 98  & 100 & 48  \\
\textbf{BBoE} & \textbf{100} & \textbf{100} & \textbf{100} & \textbf{100} & \textbf{100} & \textbf{100} & \textbf{96} & \textbf{96} \\
\bottomrule
\end{tabular}
\end{table}
\subsubsection{Cross-System Evaluation}
As seen in Fig.~\ref{fig:cross_system_results} and Table \ref{tab:success_rates}, BBoE achieves the lowest mean planning time on three of the four systems and is competitive on the fourth. It reaches 0.23 s on the unicycle ($\sim$2$\times$ faster than the next-best Db-RRT-C at 0.47 s), 0.71 s on the CWT (vs.\ 1.05 s for Db-RRT-C), and 1.66 s on the 6D quadrotor in a narrow-passage environment (vs.\ 2.07 s for RRT and 3.07 s for Db-RRT-C). On the treaded vehicle, Db-RRT-C is faster (1.23 s vs.\ 1.47 s for BBoE), though all bidirectional planners on this system cluster in the 1.2--2.0 s range. Success rates follow a similar pattern: BBoE reaches 100\% on the unicycle and CWT, and 96\% on the treaded vehicle and quadrotor. The original BoE planner fails entirely on the CWT and quadrotor and reaches only 48\% on the unicycle, demonstrating that edge reuse alone does not generalize to higher-dimensional or more constrained dynamics. BBoE produces the lowest-cost paths on the quadrotor (94.50, vs.\ 98.77--143 for the baselines) and competitive costs on the unicycle and treaded vehicle. Taken together, BBoE is the only planner that simultaneously achieves $\geq 96\%$ success on every system tested and the lowest planning time on three of four systems, spanning 3D to 6D state spaces and a range of dynamic constraints.
\subsubsection{Four-Tier Difficulty Environments}
To examine how planner performance scales with obstacle clutter, we evaluate all baselines on the unicycle system across four difficulty tiers (easy, medium, hard, very hard), with 50 random seeds per tier (Fig.~\ref{fig:unicycle_curated_results} and Table \ref{tab:success_rates}). On easy environments, all planners reach 100\% success, and BBoE is competitive with the fastest baselines (0.10 s, vs.\ 0.04 s for Db-RRT-F and 0.05 s for Db-RRT-C). The picture shifts as difficulty increases. BBoE becomes the fastest planner from medium onward, achieving 0.13 s on medium, 0.18 s on hard, and 0.35 s on very-hard, while maintaining 100\% success across all three tiers. The advantage widens with clutter: on very-hard environments, BBoE is roughly $1.4\times$ faster than the next-best GBRRT-EB at 0.48 s and substantially more reliable than the Db-RRT family, whose success rates collapse from 100\% on easy to 52\% (Db-RRT-F) and 70\% (Db-RRT-C) on very-hard. Excluding BoE, path costs remain within 10--15\% of the lowest-cost feasible planner across tiers. These results indicate that BBoE's relative advantage grows with environmental difficulty.
\subsection{Ablation Studies}
\label{sec:ablations}
We isolate BBoE's components on the unicycle, aggregating 50 seeds across all four difficulty tiers (Table~\ref{tab:bboe_ablations}). The edge-skipping parameter $N$ (Table~\ref{tab:bboe_ablations}a) is robust across two orders of magnitude: success is 50/50 for $N=5$, $20$ and $100$ and 48/50 for $N=1$, while planning time is lowest for $N=20$ and $100$ (0.46 s and 0.49 s) and about 0.7 s for $N=1$ and $N=5$. $k_\text{\upshape bias}=0.85$ and $\theta$ were fixed across systems. $\theta$ is a radius on the kinodynamic components only, set to one tenth of each component's bound on every system ($0.05$ m/s for the quadrotor), while for the unicycle, the kinodynamic subspace is empty and every edge is a neighbor. $(\delta_{\text{hr}},\gamma,k_\text{\upshape exploit})$ are inherited from GBRRT \cite{GBRRT} and shared with that baseline on each system: $(7,14,0.8)$, $(8,7,0.7)$, $(10,8,0.7)$ and $(6,6,0.95)$ for the unicycle, treaded vehicle, CWT and quadrotor. 
\begin{table*}[!t]
\centering

\tiny
\setlength{\tabcolsep}{3pt}
\renewcommand{\arraystretch}{1.1}
\caption{BBoE ablations on the unicycle environments. Time/cost: mean [min, max] over successful trials.}
\label{tab:bboe_ablations}
\begin{tabular}{@{}c@{\hspace{15pt}}c@{\hspace{15pt}}c@{}}

\begin{tabular}{@{}rcll@{}}
\multicolumn{4}{c}{(a) \texttt{num\_skip\_edges}} \\
\toprule
skip & S & T (s) & Cost \\
\midrule
1   & 48/50 & 0.69 [0.26, 2.37] & 184 [158, 213] \\
5   & 50/50 & 0.72 [0.31, 2.23] & 198 [169, 246] \\
20  & 50/50 & 0.46 [0.14, 1.48] & 201 [168, 284] \\
100 & 50/50 & 0.49 [0.16, 1.26] & 205 [160, 246] \\
\bottomrule
\end{tabular}
&
\begin{tabular}{@{}rcll@{}}
\multicolumn{4}{c}{(b) $k_{\text{bias}}$} \\[2pt]
\toprule
$k_{\text{bias}}$ & S & T (s) & Cost \\[1pt]
\midrule
0.00 & 42/50 & 2.53 [0.75, 4.97] & 453 [320, 712] \\
0.25 & 50/50 & 0.81 [0.44, 1.85] & 280 [199, 441] \\
0.50 & 50/50 & 0.69 [0.31, 1.55] & 222 [182, 329] \\
0.75 & 50/50 & 0.65 [0.28, 1.21] & 205 [169, 260] \\
1.00 & 50/50 & 0.67 [0.26, 1.43] & 192 [168, 222] \\
\bottomrule
\end{tabular}
&
\begin{tabular}{@{}lcll@{}}
\multicolumn{4}{c}{(c) Unidirectional BBoE} \\
\toprule
Difficulty & S & T (s) & Cost \\
\midrule
Easy      & 50/50 & 0.16 [0.03, 0.46] & 143 [123, 169] \\
Medium    & 50/50 & 0.12 [0.03, 0.41] & 166 [134, 225] \\
Hard      & 50/50 & 0.18 [0.04, 0.83] & 167 [132, 253] \\
Very Hard & 44/50 & 0.28 [0.03, 1.45] & 193 [147, 297] \\
\bottomrule
\end{tabular}

\end{tabular}
\end{table*}
$N$ was not fixed. The benchmarks of Sec.~\ref{sec:results} use $N=5$ for the unicycle, treaded vehicle and CWT (50/50 in Table~\ref{tab:bboe_ablations}a, though slower than $N=20$) and $N=1$ for the quadrotor, where a node has only about 40 neighboring edges against 190 for the unicycle. The biased-sampling parameter $k_{\text{bias}}$ (Table~\ref{tab:bboe_ablations}b) is flat above $0.25$ (50/50 throughout). The only drop is at $k_{\text{bias}}=0$, where pure random sampling achieves 42/50 at 2.53 s versus 50/50 at 0.65 s for $k_{\text{bias}}=0.75$, a $\sim$4$\times$ speedup confirming that goal-directed edge selection, not edge reuse alone, drives BBoE's performance. Finally, a unidirectional BBoE variant (Table~\ref{tab:bboe_ablations}c) is very fast on easy--hard environments (0.12 s--0.18 s, 100\% success) but degrades to 44/50 on very-hard, whereas bidirectional BBoE maintains 100\% success across all tiers. Thus, bidirectional search contributes negligibly in low clutter but becomes essential as obstacle density forces single-tree expansion through narrow passages.

\subsection{Discussion and Limitations}
\label{sec:limitations}
BBoE is not asymptotically optimal. For a fixed edge bundle, the states reachable from $x_0$ by sequences of neighboring edges form a countable, bundle-induced graph, and BBoE is probabilistically complete with respect to it. Every node and each of its untried neighboring edges is selected with nonzero probability, and no edge is retried from a node. Completeness over the state space holds only in the bundle-density limit \cite{BoE}. BBoE performs no iterative cost descent and cannot be trapped in a local minimum in the optimization sense. Biased edges can stall when all of $\boldsymbol{\beta}_{\text{\upshape biased}}$ collides, which is why $\boldsymbol{\beta}_{\text{\upshape random}}$ exists. It fails on a \textsl{coverage gap} (i.e., when no sequence of edges threads the free space), an ill-chosen $\theta$, dynamics not invariant under the rigid-body edge transformation, or bundle memory and neighbor-query cost, which grow with the kinodynamic dimension (Table~\ref{tab:system_specs} holds 190 to 100\,000 edges, and a node has 190, 330, 22, and 40 neighbors for the unicycle, treaded vehicle, CWT and quadrotor). We confirmed the coverage-gap case on a unicycle environment whose only passage requires a 90° turn within 0.12 m of free width, while the tightest turning radius in the bundle is 0.64 m. Here, BBoE fails on all 50 seeds at 5 s and on all 10 seeds at 30 s, although a continuous-control solution exists.

\section{CONCLUSION AND FUTURE WORK}

In this work, we presented BBoE, a bidirectional kinodynamic motion planner that leverages a precomputed bundle of edges to outperform existing baselines across various systems and environments. BBoE is the only evaluated planner to achieve $\geq$96\% success across every system tested while attaining the lowest planning time on three of four, with its advantage growing alongside obstacle density. Using lazy search tactics with the edge sequencing strategy to optimize propagation in clutter, optimizing travel time, and replacing the hand-tuned $k_\text{\upshape bias}$ and $\text{N}$ with adaptive or learning-based selection, represent promising directions for future work.

\bibliographystyle{IEEEtran}
\bibliography{reference_final.bib}
\end{document}